\documentclass[runningheads]{llncs}

\usepackage{eccv}

\usepackage{eccvabbrv}

\usepackage{graphicx}
\usepackage{booktabs}
\usepackage{soul}
\usepackage{float}
\usepackage{makecell}
\usepackage{graphicx}
\usepackage{marvosym}
\newcommand{\equalcontrib}{\textsuperscript{\textdagger}}
\newcommand{\corrauthor}{\textsuperscript{\Letter}}
\newcommand{\internmark}{\textsuperscript{*}}

\usepackage[accsupp]{axessibility}  % Improves PDF readability for those with disabilities.

\usepackage{hyperref}

\usepackage{orcidlink}
\usepackage{pifont}
\usepackage{booktabs}
\usepackage{multirow}
\usepackage{amsmath}

\begin{document}

% \title{MOJITO: Multi-Modal Joint Interaction for \\ Trajectory Optimization} 
\title{MOJITO: Modal Joint Learning for Unified \\ End-to-End Autonomous Driving}

\titlerunning{MOJITO}

\author{
Zhijing Cheng\inst{1,2}\internmark\equalcontrib \and
Xuancheng Zhang\inst{2}\equalcontrib \and
Donglin Di\inst{2} \and
Lei Fan\inst{3} \and 
Baorui Ma\inst{2} \and 
Hao Li\inst{2} \and 
Xun Yang\inst{1}\corrauthor
}

\authorrunning{Z. Cheng et al.}

\institute{
University of Science and Technology of China, China \and
Li Auto, China \and University of New South Wales, Australia \\
\email{mumucc@mail.ustc.edu.cn, xyang21@ustc.edu.cn, \{zhangxuancheng,didonglin,mabaorui,lihao43\}@lixiang.com}
}

\maketitle

\begingroup
\renewcommand{\thefootnote}{}
\footnotetext{\textsuperscript{*}
Work done during an internship at Li Auto.}
\footnotetext{\textsuperscript{\textdagger} Equal contribution.}
\footnotetext{\textsuperscript{\corrauthor} Corresponding author.}
\endgroup

\newcommand{\red}[1]{{\color{red}#1}}
\newcommand{\todo}[1]{{\color{red}#1}}
\newcommand{\TODO}[1]{\textbf{\color{red}[TODO: #1]}}

\newcommand{\czj}[1]{\textcolor{red}{#1}}

% \begin{abstract}
%   The abstract should concisely summarize the contents of the paper. 
%   While there is no fixed length restriction for the abstract, it is recommended to limit your abstract to approximately 150 words.
%   Please include keywords as in the example below. 
%   This is required for papers in LNCS proceedings.
%   \keywords{First keyword \and Second keyword \and Third keyword}
% \end{abstract}

\begin{abstract} End-to-end autonomous driving systems commonly follow a cascaded two-stage pipeline where a perception stage compresses multi-modal sensor inputs into a compact context and a downstream planner predicts trajectories conditioned on this context.
%We argue that this one-way interface introduces an information bottleneck that discards fine-grained details which are critical for planning and often requires auxiliary supervision or predefined anchors. Moreover, this separation limits the planner from  leveraging the full representational power of modern vision foundation models, because the planner only receives a compressed context that cannot retain such expressive features.
We argue that this one-way perception-to-planning interface forces sensor inputs into a compact representation, losing the fine-grained details critical for planning. Moreover, by constraining the planner to this compressed context, it is difficult to leverage the rich representations offered by modern vision foundation models.
To address these issues, we propose \textbf{MOJITO}, a unified sensor-to-action framework for end-to-end autonomous driving built on modal joint learning. MOJITO removes the cascaded interface and instead performs block-wise Modal Joint Attention that simultaneously updates action, image, and LiDAR features, allowing the planner to directly access  multi-modal features during  action generation. 
% The model uses Transformer encoders for vision and LiDAR, a pillar-based LiDAR tokenizer, and an anchor-free diffusion planner implemented with a Diffusion Transformer.
% To address these issues, we propose \textbf{MOJITO} (\textbf{MO}dal \textbf{J}o\textbf{I}nt learning for end to end au\textbf{TO}nomous driving), a unified sensor to action framework for end-to-end autonomous driving with three parallel branches, including a vision branch built on ViT, a LiDAR branch following a ViT style design with a PillarGroup tokenizer, and an action planner based on a Diffusion Transformer (DiT). Crucially, we introduce Modal Joint Attention that enables action tokens to interact with dense sensor tokens at each block, so the planner can directly attend to fine-grained multi modal feature during trajectory generation. 
% This deep cross modal interaction naturally constrains the diffusion process to feasible regions, removing the reliance on anchors and heavy auxiliary supervision while maintaining compatibility with Transformer based foundation models.
% Without anchors or auxiliary supervised tasks.
MOJITO achieves 88.9 PDMS on the NAVSIM v1 dataset and 88.4 EPDMS on the more challenging NAVSIM v2 dataset, setting a new state-of-the-art. Extensive experiments further demonstrate strong scalability, instruction following, and diverse trajectory generation. Code and models are
available at \url{https://github.com/mumucc01/MOJITO}.
\end{abstract} 
\section{Introduction}
Autonomous driving has witnessed significant progress with the evolution of deep learning~\cite{bojarski2016end2end,liao2025diffusiondrive,hu2022st,sun2024sparsedrive,driverwm}. A long-standing goal is to map sensory observations directly to driving outputs, motivating increasing interest in end-to-end models. Unlike traditional modular stacks, these approaches optimize perception and planning jointly. A core challenge is bridging low-level perception with high-level decision-making. To address this, most methods adopt a cascaded two-stage pipeline. A perception stage extracts  intermediate context, and a planning stage generates the action based on it. 
Typically, this relies on specific encoders like CNNs~\cite{hu2023uniad, chitta2022transfuser, liao2025diffusiondrive} to produce compact representations~\cite{huang2022bevdet4d}. More recently, Vision-Language-Action (VLA) models~\cite{tian2025drivevlm, zhou2025autovla,li2025recogdrive,wang2025omnidrive} have advanced this stage by leveraging the rich semantic reasoning of Vision-Language Models (VLMs). However, these advanced methods still compress vision-language information into coarse  tokens for planning. While semantically enriched,  this compression sacrifices the fine-grained geometric precision required for trajectory prediction and  is  processed by separate heads like MLPs~\cite{li2025spacedrive} or diffusion models~\cite{liao2025diffusiondrive, jiang2025diffvla}.

\begin{figure}[t]
\centering
\includegraphics[width=0.73\linewidth]{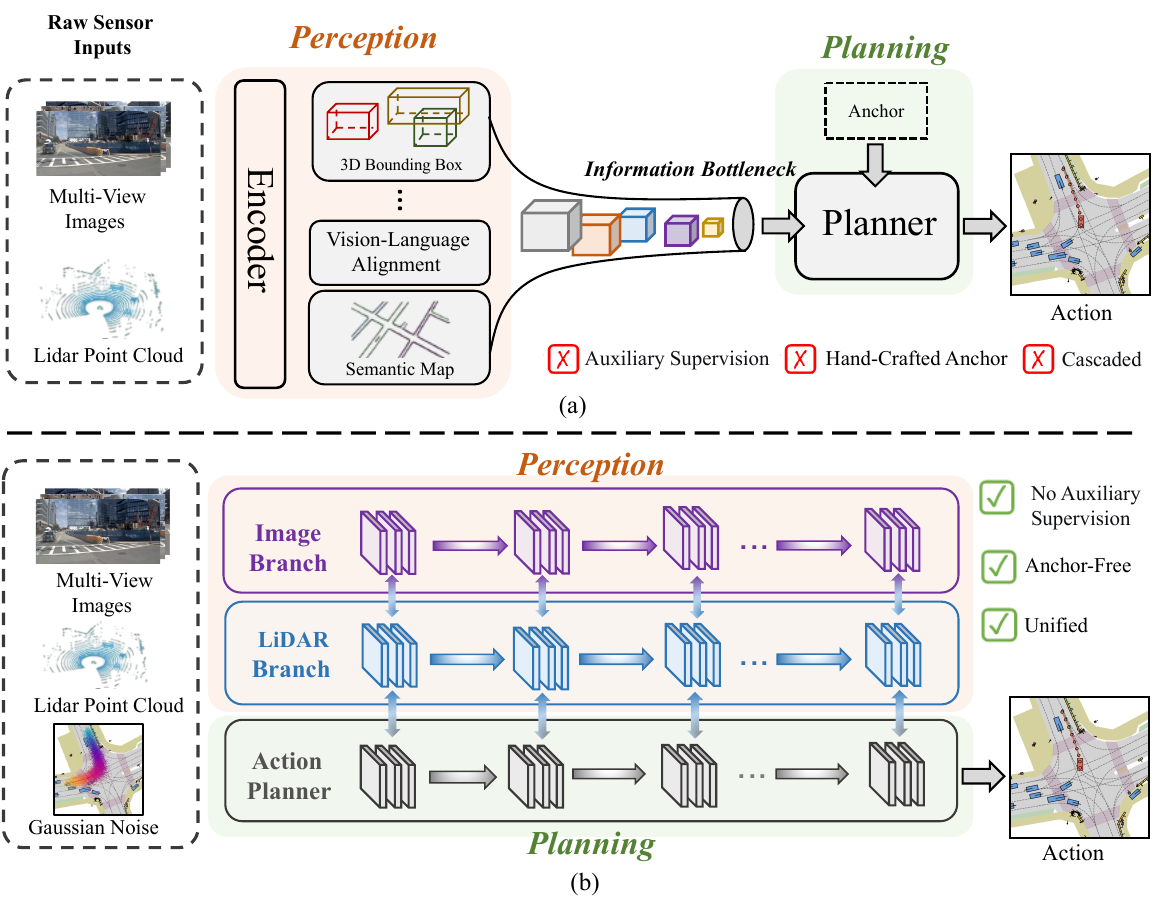}
\vspace{-0.5em}
\caption{This figure outlines the core differences between a cascaded two-stage framework and our unified E2E architecture. (a) shows that the  ``perception-to-planning'' methods  compress the fine-grained sensor information and suffer from the information bottleneck. (b) We propose a unified multimodal architecture comprising three parallel branches. Benefiting from the multi-modal fusion, MOJITO can generate  feasible and stable actions without relying on predefined anchors or auxiliary supervision.}
\vspace{-1em}
\label{teaser}
\end{figure}

This two-stage design limits the potential of end-to-end driving in three ways. First, as shown in Fig. \ref{teaser} (a), it imposes a one-way ``perception-to-planning'' interface which  brings an information bottleneck. This compression process discards fine-grained details, resulting in the planner lacking access to the sensor data for decision-making. 
Second, relying exclusively on latent context necessitates manual constraints, which require costly annotations and limit scalability. As the planner is conditioned on compressed representations, the supervision from the trajectory is too weak to guide the context representation learning. To compensate, methods require auxiliary tasks like 3D box prediction~\cite{zheng2025diffusionplanner,chitta2022transfuser,li2019gs3d} or introduce predefined 
% \czj{high-cost} 
anchors~\cite{liao2025diffusiondrive,li2025gtrs} to stabilize generation.
Third, this structure creates a compatibility gap. Modern foundation models are built on standard Transformers~\cite{simeoni2025dinov3,qwen3}, yet existing methods restrict them to the role of perception encoders~\cite{zhou2025autovla}. This prevents the planner from leveraging the full expressive power of Vision Transformers (ViTs). 
As a result, the rich representation ability from pre-trained weights is not fully utilized for decision-making.

Motivated by these limitations, we aim to build a direct ``vision-to-action''  framework that avoids the information bottleneck and heavy auxiliary supervision while maintaining compatibility with vision foundation models. To this end, we propose MOJITO (\textbf{MO}dal \textbf{J}o\textbf{I}nt learning for unified end-to-end 
au\textbf{TO}nomous driving).  
As illustrated in Fig. \ref{teaser} (b), our framework consists of three parallel branches: an Image branch (ViT), a LiDAR branch following a ViT-style design that utilizes a PillarGroup tokenizer, and an Action Planner branch built on a Diffusion Transformer (DiT)~\cite{peebles2023dit}. Crucially, instead of isolating the planner behind a compressed latent context, we introduce \textbf{Modal Joint Attention}. This mechanism allows action tokens to interact with dense sensor tokens at each block, enabling the planner to directly attend to fine-grained visual details. This deep interaction naturally constrains the diffusion process to feasible regions, removing the need for hand-crafted anchors or auxiliary supervised tasks.

To validate our approach, we evaluate MOJITO on the \textbf{NAVSIM} benchmark. Without relying on predefined anchors or heavy auxiliary supervision, our unified architecture achieves \textbf{88.9 PDMS} on the NAVSIM-V1 \textit{navtest} split, significantly outperforming recent strong E2E and VLA baselines such as DiffusionDrive~\cite{liao2025diffusiondrive} (88.1 PDMS) and ReCogDrive~\cite{li2025recogdrive} (86.5 PDMS). Furthermore, MOJITO establishes a new state-of-the-art performance  (\textbf{88.4 EPDMS}) on the more challenging NAVSIM-V2 \textit{navtest} split. Beyond standard metrics, extensive experiments also highlight MOJITO's scalability, instruction following, and diverse trajectory generation capabilities. These results demonstrate that directly bridging raw multi-modal sensor inputs with the action planner not only simplifies the architectural pipeline but also effectively improves planning quality and driving stability in complex scenarios.

The main contributions of this work are as follows:
\begin{itemize}
    \item We propose MOJITO, a fully unified end-to-end driving framework that integrates multi-modal sensors with a diffusion-based planner via a novel Modal Joint Attention mechanism. This design resolves the information bottleneck caused by cascaded pipelines by allowing action tokens to interact deeply with dense sensor features at each block.

    \item We develop a simplified, anchor-free architecture that removes the dependency on auxiliary supervision and predefined anchors. By following standard ViT and DiT design protocols, our approach maintains full compatibility with foundation models, allowing the planner to utilize pre-trained representations directly for decision-making.

    \item We achieve leading performance on both NAVSIM-v1 and v2 benchmarks, surpassing strong existing baselines. Furthermore, extensive experiments de-monstrate MOJITO's robust capabilities in scalability, instruction following, and diverse trajectory generation.

\end{itemize}

\section{Related Work}
\label{sec:related_work}
\vspace{-0.5em}
\subsection{Architectures for Autonomous Driving}
\vspace{-0.2em}
 Recent years have witnessed significant progress in end-to-end autonomous driving, moving from modular pipelines to integrated systems~\cite{li2024bevformer,liao2025diffusiondrive,lu2025uniugp,su2026drivemamba}. Pioneering frameworks like UniAD~\cite{hu2023uniad} defined the standard E2E paradigm, where perception features are extracted  and then fed into a planning head. Following this, works like SparseDrive~\cite{sun2024sparsedrive} and HydraMDP~\cite{li2024hydra} refined feature representations but maintained this two-stage design. Subsequently, generative approaches such as DiffusionDrive~\cite{liao2025diffusiondrive} and GTRS~\cite{li2025gtrs}  replaced conventional planners with diffusion models, but still relied on perception features as conditions and predefined trajectory anchors for stable generation. 
 In parallel, recent advances in multimodal learning have demonstrated the potential of MLLMs for visual language understanding, semantic reasoning, and robust visual representation learning~\cite{wang2026affordbot,ijcvlearning,yang2026fine,yang2021deconfounded}.
 Inspired by these advances, Vision-Language-Action (VLA) methods ~\cite{li2025spacedrive, jiang2025diffvla,gaussiandwm} enhance perception through semantic reasoning, yet continue to pass compressed embeddings to a separate planner.
Overall, whether utilizing traditional MLPs, diffusion processes, or VLA architectures, the cascaded design enforces a one-way information flow. This structural separation limits the mutual interaction between scene understanding and action planning, motivating our fully unified, tokenized approach.

\subsection{Diffusion Models for Planning}
Diffusion models have shown strong capability in generating diverse and high-quality samples across various domains \cite{wan2025wan,everybodydance,moca}, motivating researchers to transfer this strength to autonomous driving  to improve trajectory diversity and enable capabilities like instruction following and driving style adaptation~\cite{wang2025diffad,tu2024driveditfit,yang2024diffusiones,liao2025diffusiondrive}. Motion Diffuser~\cite{jiang2023motiondiffuser} represents an early effort in applying diffusion models to action generation, learning a joint distribution of multi-agent trajectories and enabling diverse future predictions without predefined anchors. Diffusion Planner~\cite{zheng2025diffusionplanner} extends diffusion modeling to closed-loop planning, producing diverse ego trajectories and supporting joint prediction-planning tasks. Another approach, Diffusion ES~\cite{yang2024diffusiones}, validated the importance of trajectory diversity by achieving instruction-following behaviors through optimization with black-box rewards, even generating novel actions not present in the training data. Despite these advancements, most diffusion-based planners still rely on vectorized environmental representations and cannot directly process sensor data, which restricts their integration into end-to-end driving.

\subsection{Unified Multimodal Architectures}
Unified multimodal architectures aim to jointly model diverse modalities and tasks within a single framework~\cite{du2026unsupervised}. Existing methods~\cite{deng2025bagel,bi2026motus} achieve unification by sharing self-attention layers between understanding and generation experts. In autonomous driving, 
%such unified modeling is still in its early stages but is attracting significant attention. 
recent works like UniUGP~\cite{lu2025uniugp} propose a hybrid expert architecture to synergize scene understanding, future video generation, and trajectory planning. 
They essentially unify multiple generative branches (e.g., video generation), our method instead couples discriminative perception (DINOv3, Uni3D) with generative planning (DiT).
Unlike existing pipelines~\cite{drivetransformer}, our method learns optimal continuous  action representations within the sparse perceptual space, with trajectory supervision only.

\section{Problem Formulation and Challenges}

\subsection{Task Definition}
The goal of end-to-end autonomous driving is to generate a sequence of future waypoints for the ego vehicle directly from raw sensor inputs. At each planning step, the model receives multi-view camera images \(\mathcal{I} = \{I_1, I_2, \dots, I_N\}\) and a LiDAR point cloud \(\mathcal{L}\), capturing the semantic and geometric context of the scene. The output is a trajectory represented as a sequence of waypoints \(\tau = \{w_t\}_{t=1}^{T_f}\), where each waypoint \(w_t = [x_t, y_t, \theta_t]\) contains the 2D position and heading angle of the ego vehicle over a planning horizon \(T_f\) (in our case, \(T_f = 8\)). 

The task is defined as learning a mapping \(f_\theta : (\mathcal{I}, \mathcal{L}) \rightarrow \tau\) that predicts smooth and feasible trajectories from raw visual inputs. In our framework, this mapping is realized through a diffusion process, enabling flexible, data-driven trajectory generation without hand-crafted anchors or auxiliary supervised tasks.

\subsection{Challenges in End-to-End Planning}
While recent end-to-end driving models have shown promising results, unifying multi-modal perception and trajectory planning remains difficult due to two primary challenges:

\textbf{Challenge 1: Unifying Perception and Planning Architectures.}  Most existing end-to-end frameworks \cite{chitta2022transfuser,liao2025diffusiondrive} treat perception and planning as isolated stages connected by a bottleneck. They typically employ CNN-based backbones (like ResNet34) to extract a compact ``context feature'' map, which is then passed to a separate planning module as a condition. This paradigm prevents the fine-grained feature-level interaction between the planned trajectory and the perceived environment. The planner only passively receives the compressed context, rather than actively querying the sensor information during the action planning. A unified architecture that allows bidirectional information flow between perception and planning is required.

\begin{figure}[t]
%\centering
\hspace{1cm}
\includegraphics[width=0.8\linewidth]{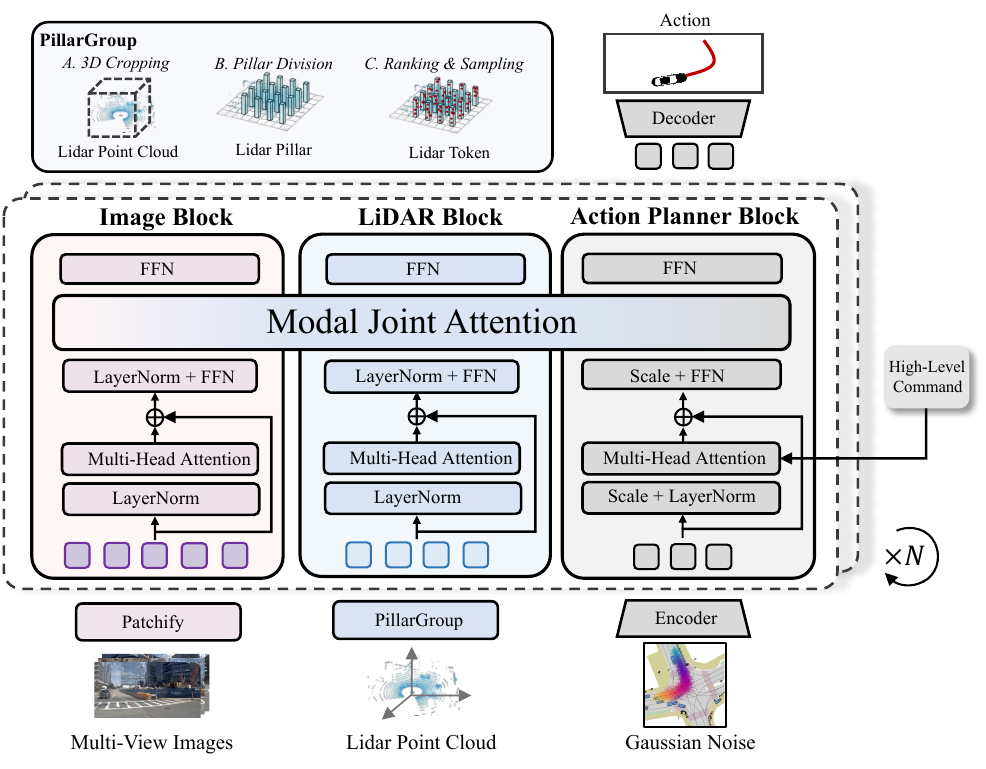}
\caption{MOJITO Architecture. To avoid the information bottleneck  inherent in  cascaded frameworks, we propose MOJITO, a unified multi-modal framework with three branches: Image, LiDAR, and the Action Planner branch. To preserve the 3D spatial understanding of the driving scene, we  introduce a BEV Grid Patchify module named PillarGroup. The Modal Joint Attention mechanism enables  
the deep interaction between the planned trajectory and the perceived environment, achieving a bidirectional flow of rich geometric and semantic information during the generation process.}
\label{pipeline}
\vspace{-0.5em}
\end{figure}

\textbf{Challenge 2: Preserving Multi-Modal Information.} 
Autonomous driving relies on heterogeneous data: images provide dense semantic information, LiDAR provides sparse but accurate 3D geometry~\cite{deng2025best}, and trajectories are 2D sequential coordinates. Previous methods often force these modalities into a single format, such as projecting 3D point clouds into 2D Bird's-Eye-View (BEV) images. This rough transformation discards fine-grained geometric details and height information. Meanwhile, point-based methods like Farthest Point Sampling (FPS) \cite{eldar1997fps} and K-Nearest Neighbors (KNN) \cite{guo2003knn,zhang2017knn2}, commonly used in shape-level tasks, struggle in large-scale driving scenes because they lack absolute physical scale. 
Moreover, they struggle with uneven LiDAR density, splitting dense objects across multiple
tokens while merging sparse ones. This instability degrades
multi-modal alignment.
A tokenization strategy that preserves the absolute metric scale of LiDAR is essential for effective multi-modal fusion.

\section{Method}
To address the aforementioned challenges, we propose a fully unified Transformer-based architecture that integrates multi-view images, LiDAR point clouds, and action planning into a single framework.  
%Unlike cascaded designs where perception and planning are isolated into separate stages, our framework achieves the deep interaction between the planned trajectory and the perceived environment.  This simplified architecture allows the planner to ..., achieving the leading performance without relying on hand-crafted anchors and heavy auxiliary supervision.

\subsection{Overall Architecture}
As illustrated in Fig. \ref{pipeline}, our model consists of three parallel branches: the Image Branch, the LiDAR Branch, and the Action Planner. 
These three branches adopt structurally identical but independent Transformer architectures, each comprising 12 blocks with a hidden state dimension of 384. 

Instead of extracting perception features first and passing them to the planner, our architecture processes all modalities simultaneously. The core of this design is the \textbf{Modal Joint Attention} mechanism. In each block, tokens from the image, LiDAR, and  action planner branches are concatenated and processed through a shared self-attention layer. This allows every modality to directly interact with the others at every depth of the network. Moreover, to maintain spatial and sequential awareness across this mixed-modality sequence, a learnable positional embedding is dynamically generated and added within each layer.

\subsection{Perception Branches}
To enable the Modal Joint Attention, we first convert the raw sensor inputs into sequences of tokens with a unified dimension using dedicated perception branches.

\textbf{Image Branch:} 
The image feature is processed using a DINOv3-S+ architecture. For the initial tokenization, each multi-view image is divided into \(16 \times 16\) patches. These patches are linearly projected into image tokens before being fed into the Transformer blocks to extract dense semantic features.

\textbf{LiDAR Branch:}
For the LiDAR branch, we adopt the Uni3D-S architecture. Uni3D was originally designed for shape-level representation, relying on Farthest Point Sampling (FPS) and K-Nearest Neighbors (KNN) for tokenization. We observe that this tokenization design does not align well with the requirements of autonomous driving (AD) scenarios. FPS and KNN operate on relative spatial distances, discarding the absolute metric scale and global geometric layout required for scene-level navigation.

Inspired by grid-based BEV representations~\cite{li2024bevformer,huang2022bevdet4d}, we introduce a BEV Grid Patchify module named \textbf{PillarGroup} for LiDAR tokenization. 
PillarGroup crops the 3D point cloud space and divides the space into a fixed grid (e.g., \(32 \times 32\) pillars), covering a specific metric range. Points are assigned to their corresponding physical pillars. We select the top-\(K\) non-empty pillars (\(K=512\), padded if necessary) and sample a fixed number \(N\) of points per pillar (\(N=64\)). Because each pillar covers a deterministic physical area, the generated tokens possess consistent semantic information and absolute physical coordinates.

\subsection{Action Planner}
The action planner generates future trajectories through a diffusion process parameterized directly within our unified Transformer architecture. Unlike traditional pipelines that extract perception features first and apply them as conditions, our approach treats action waypoints as native tokens that evolve jointly with perception tokens.

Formally, the ground-truth future trajectory is represented as:
 
\begin{equation}
\tau^{(0)} = [x_{\text{ego}}(1), x_{\text{ego}}(2), \dots, x_{\text{ego}}(T)],
\end{equation}
where each state \(x_{\text{ego}}(t)\) encodes the ego vehicle's position and heading as \(x_{\text{ego}}(t) = [x_t,\, y_t,\, \theta_t]\), and \(T\) is the predicted steps. During inference, the entire trajectory sequence is initialized as pure Gaussian noise. A forward process progressively corrupts the ground-truth states with noise, yielding a sequence of noisy trajectories \(\tau^{(k)}\) at diffusion step \(k\).

During training, our unified network \(\epsilon_\theta\) learns to recover the data distribution from noisy data with the following objective:
\begin{equation}
\mathcal{L}_{\theta} = \mathbb{E}_{x^{(0)}, k \sim U(0,1), x^{k} \sim q_{k0}(x^{(k)} | x^{(0)})} \left[ \|\epsilon_{\theta}(x^{(k)}, k, C) - x^{(0)}\|^2 \right].
\end{equation}

The diffusion step \(k\) is embedded via an MLP, combined with the high-level command (e.g., turn left and go straight), and injected into the action tokens using Adaptive Layer Normalization (adaLN). \(C\) consists of the perception tokens and the high-level command. The perception features and the action tokens are fused naturally through our modal joint attention mechanism.

% \subsection{Unified Joint Attention}
% To enable deep interaction across modalities, we introduce the Unified Joint Attention mechanism. Let \(X_I\), \(X_L\), and \(X_A\) denote the token sequences from the image, LiDAR, and action branches, respectively. In each Transformer block, we concatenate them along the sequence dimension to form a unified representation:
% $X_{joint} = [X_I \parallel X_L \parallel X_A]$.
% After adding modality-specific positional embeddings, \(X_{joint}\) is processed through a shared Multi-Head Self-Attention (MHSA) layer. 

\subsection{Modal Joint Attention}
To enable interaction across modalities, we introduce the Modal Joint Attention mechanism within each Transformer block. Let \(X_I \in \mathbb{R}^{N_I \times D}\), \(X_L \in \mathbb{R}^{N_L \times D}\), and \(X_A \in \mathbb{R}^{N_A \times D}\) denote the token sequences from the image, LiDAR, and action branches, respectively, where \(D = 384\) is the hidden dimension.

Within each block, we concatenate the tokens along the sequence dimension to form a unified representation:
\begin{equation}
X_{joint} = [X_I \parallel X_L \parallel X_A] \in \mathbb{R}^{(N_I + N_L + N_A) \times D}.
\end{equation}

We add learnable positional embeddings to \(X_{joint}\) to retain spatial and sequential context. The joint sequence is then processed through a shared Multi-Head Self-Attention (MHSA) layer:
\begin{equation}
Y_{joint} = \text{MHSA}(\text{LN}(X_{joint} + PE)) + X_{joint},
\end{equation}
where the \(\text{PE}\) denotes the position embeddings, and the \(\text{LN}\) denotes Layer Normalization. This operation allows the action tokens to query geometric and semantic features directly from the perception tokens, while the perception branches adjust their representations based on the planning context. Finally, \(Y_{joint}\) is split back into the updated tokens \(Y_I\), \(Y_L\), and \(Y_A\), which are passed to their respective Feed-Forward Networks (FFN).

% \begin{figure}[tbp]
% \centering
% \includegraphics[width=0.9\linewidth]{Section/picture/attn.pdf}
% \caption{The overview of attn map.
% }
% \label{pipeline}
% \end{figure}

This design embodies our core modality fusion philosophy. It preserves the distinct functional roles of perception and action experts without introducing task interference, while enabling effective cross-modal feature fusion. By projecting all modalities into a shared representational space, diverse pretrained knowledge from the vision and 3D domains naturally complements the action generation. During the self-attention operation, action tokens directly query the fine-grained geometric and semantic information from  perception tokens, while the perception features adaptively update based on the planning intent. Finally, the updated joint sequence is split back into \(X_I\), \(X_L\), and \(X_A\) as the inputs of the subsequent block.

\section{Experiment}
\subsection{Dataset}

% \textbf{NAVSIM.} The NAVSIM ~\cite{Dauner2024navsim} is a large-scale dataset tailored for motion planning research, derived from the OpenScene~\cite{openscene2023} and nuPlan~\cite{caesar2021nuplan} frameworks. It provides a diverse collection of real-world driving sequences that emphasize decision-intensive scenarios rather than routine cruising. Each sample contains synchronized data from eight cameras offering 360° coverage and a fused LiDAR point cloud aggregated from five sensors, accompanied by 2\,Hz. We evaluate planning performance under a \textbf{non-reactive closed-loop} setting, where the ego trajectory is executed within a simulation environment but without environment feedback or reactive agents. The benchmark employs the \textbf{Planning Deviation Metric Score (PDMS)}~\cite{Dauner2024navsim}, a weighted composite measure consisting of several sub-metrics: no at-fault collisions (NC), drivable area compliance (DAC), time-to-collision (TTC), comfort (Comf.), and ego progress (EP).

\textbf{NAVSIM.} NAVSIM~\cite{Dauner2024navsim} is a large-scale dataset for trajectory planning, based on OpenScene~\cite{openscene2023} and nuPlan~\cite{caesar2021nuplan}. It provides real-world driving sequences focused on decision-heavy scenarios, featuring synchronized 360° camera and LiDAR data sampled at 2\,Hz. We evaluate planning in a non-reactive closed-loop setting, where the ego trajectory is simulated without environment feedback. 

% \textbf{NAVSIM-v1 and v2.} 
The original NAVSIM-v1 evaluates planners using the \textbf{Predictive Driver Model Score (PDMS)}~\cite{Dauner2024navsim}, which combines metrics like no at-fault collisions (NC), drivable area compliance (DAC), time-to-collision (TTC), comfort, and ego progress (EP).  The NAVSIM-v2 update introduces a stricter evaluation with the \textbf{Extended PDMS (EPDMS)}, adding finer compliance checks such as Driving Direction Compliance (DDC), Traffic Light Compliance (TLC), Lane Keeping (LK), History Comfort (HC), and Extended Comfort (EC). 
% Additionally, v2 introduces the \textbf{Navhard} split, which provides these synthetic futures and stresses robustness in more challenging scenarios.  NAVSIM-v2 further defines a \textbf{two-stage} protocol: Stage-1 evaluates the planned trajectory on logged sequences, while Stage-2 uses additional future annotations (available in \textit{Navhard}) to evaluate planning under a stronger setting.
NAVSIM-v2 additionally introduces the \textbf{\textit{navhard}} split, which provides synthetic futures and stresses robustness in more challenging scenarios. 
Moreover, NAVSIM-v2 further defines a \textbf{two-stage} protocol: Stage-1 evaluates the planned trajectory on logged sequences, while Stage-2 leverages the synthetic follow-up observations to evaluate the planner under a stronger setting.
% \subsection{Implementation Detail}
% We choose a pretrained DINOv3-S+~\cite{simeoni2025dinov3} as our image branch. We fine-tune it to adapt its dense visual representations to driving-specific semantics such as road layout, traffic objects, and scene geometry.
% Meanwhile, we employ the Uni3D-S~\cite{zhou2023uni3d} backbone for the LiDAR branch, which adopts a ViT-style architecture for structured 3D feature extraction. The action branch is modeled as an anchor-free diffusion process in noise space, which is implemented with a Diffusion Transformer~\cite{peebles2023dit}.
% During the training and inference, we use three cropped and downscaled forward-facing camera images, concatenated as a 1024×256 image, and the raw lidar points. Our method is trained on navtrain split from scratch for 200 epochs with AdamW optimizer on 8 NVIDIA H200 GPUs with total batch size of 512, setting the learning rate to $6 \times 10^{-4}$ . No test-time augmentation is applied and the final output for evaluation on \textit{navtest} split is 8-waypoint trajectory over 4 seconds.
\vspace{-0.5em}
\subsection{Implementation Details}
We adopt a  DINOv3-S+~\cite{simeoni2025dinov3} as our image branch and a  Uni3D-S~\cite{zhou2023uni3d} as our LiDAR branch. We initialize both backbones from their originally released weights and do not perform extra pretraining on the autonomous-driving data. The action branch is modeled as an anchor-free diffusion process in the noise space, implemented with a Diffusion Transformer~\cite{peebles2023dit}.  During training, the image and
LiDAR branches are fine-tuned, while the action planner is trained from scratch.

During training and inference, we use three downscaled camera views (front, $60^\circ$  left, $60^\circ$ right), which are concatenated to a resolution of $1024 \times 256$, and the raw LiDAR points as input. We follow Transfuser~\cite{chitta2022transfuser} for image pre-processing.  We train MOJITO on the \textit{navtrain}  using AdamW on 8 NVIDIA H200 GPUs with a total batch size of 512 and a learning rate of \(6 \times 10^{-4}\).  For both NAVSIM-v1 and v2, we predict an 8-waypoint trajectory over 4 seconds for evaluation.

% \subsection{Implementation Detail}
% We use a pretrained DINOv3-S+~\cite{simeoni2025dinov3} as our image branch and a pretrained Uni3D-S~\cite{zhou2023uni3d} as our LiDAR branch. We initialize both backbones from their original released weights and do not perform extra pretraining on autonomous-driving data. The action branch is modeled as an anchor-free diffusion process in noise space, implemented with a Diffusion Transformer~\cite{peebles2023dit}. During training, all parameters are updated end-to-end and no modules are frozen.

% During training and inference, we use three downscaled camera views (left, front, right) and the raw LiDAR points as input. We train on the \textit{navtrain} split for 200 epochs using AdamW on 8 NVIDIA H200 GPUs with a total batch size of 512 and a learning rate of \(6 \times 10^{-4}\). No additional data augmentation or test-time preprocessing is applied. For both NAVSIM-v1 and v2, we predict an 8-waypoint trajectory over 4 seconds for evaluation.

\subsection{Quantitative Comparison on NAVSIM-v1}
% \begin{table}[]
% \caption{The Quantitative Comparision on NAVSIM-v1}
% \setlength{\tabcolsep}{1.3mm}\begin{tabular}{l|ccc|cccccc}
% \hline \toprule
% Method         & Input  & Anchor & Params. & NC   & DAC  & TTC  & Comf. & EP   & PDMS \\  \midrule
% UniAD          & Camera & 0      &         & 97.8 & 91.9 & 92.9 & 100   & 78.8 & 83.4 \\
% LTF            & Camera & 0      &         & 97.4 & 92.8 & 92.4 & 100   & 79   & 83.8 \\
% Transfuser     & C \& L & 0      &         & 97.7 & 92.8 & 92.8 & 100   & 79.0 & 84.0 \\
% DRAMA          & C \& L & 0      &         & 98.0 & 93.1 & 94.8 & 100   & 80.1 & 85.5 \\
% VADv2          & C \& L & 8192   &         & 97.2 & 89.1 & 91.6 & 100   & 76.0 & 80.9 \\
% Hydra-MDP      & C \& L & 8192   &         & 97.9 & 91.7 & 92.9 & 100   & 77.6 & 83.0 \\
% DiffusionDrive & C \& L & 20     &         & 98.2 & 96.2 & 94.7 & 100   & 82.2 & 88.1 \\
% WoTE           & C \& L & 256    &         & 98.5 & 96.8 & 81.9 & 94.9  & 99.9 & 88.3 \\
% DriveSuprim    & C \& L & 0      &         & 97.8 & 97.3 & 86.7 & 93.6  & 100  & 89.9 \\
% ReCogDrive     & Camera & 0      &         & 97.9 & 97.3 & 94.9 & 100   & 87.3 & 90.8 \\ \midrule
% Ours           & C \& L & 0      &         &      &      &      &       &      & 90.9 \\  \bottomrule
% \end{tabular}
% \label{table_quantitative_comparsion_on_navsim_v1}
% \end{table}

\begin{table}[t]
\caption{Quantitative comparison with end-to-end planning methods on the planning-oriented NAVSIM-v1 \textit{navtest} split. ``Input'' indicates sensor modalities, where ``C \& L'' denotes using both camera and LiDAR. ``Anchor'' denotes the number of anchors used by anchor-based planners (0 for anchor-free). The best results are marked in bold.}
\vspace{-0.8em}

\setlength{\tabcolsep}{0.9mm}\begin{tabular}{l|cc|cccccc}
\hline \toprule
Method         & Input  & Anchor & NC $\uparrow$  & DAC $\uparrow$ & TTC$\uparrow$  & Comf. $\uparrow$& EP $\uparrow$  & PDMS$\uparrow$ \\  \midrule
Constant Velocity         & - &  -     & 68.0 & 57.8 & 50.0 & \textbf{100}    & 19.4  & 20.6 \\
Ego Status MLP     & - & -     & 93.0 & 77.3 & 83.6 & \textbf{100}    & 62.8 & 65.6 \\  \midrule

UniAD~\cite{hu2023uniad}         & Camera & 0      & 97.8 & 91.9 & 92.9 & \textbf{100}    & 78.8 & 83.4 \\
Transfuser~\cite{chitta2022transfuser}      & C \& L & 0      & 97.7 & 92.8 & 92.8 & \textbf{100}    & 79.2 & 84.0 \\
DRAMA~\cite{yuan2024drama}          & C \& L & 0      & 98.0 & 93.1 & 94.8 & \textbf{100}    & 80.1 & 85.5 \\
VADv2~\cite{chen2024vadv2}          & C \& L & 8192   & 97.2 & 89.1 & 91.6 & \textbf{100}    & 76.0 & 80.9 \\
Hydra-MDP~\cite{li2024hydra}     & C \& L & 8192   & 98.3 & 96.0 & 94.6 & \textbf{100}    & 78.7 & 86.5 \\
DiffusionDrive~\cite{liao2025diffusiondrive} & C \& L & 20     & 98.2 & 96.2 & 94.7 & \textbf{100}    & 82.2 & 88.1 \\
WoTE~\cite{li2025wote}           & C \& L & 256    & 98.5 & 96.8 & \textbf{94.9} & \textbf{100}   & 82.6 & 88.3 \\
% DriveSuprim\cite{yao2025drivesuprim}    & C \& L & 0      & 97.8 & 97.3 & 93.6 & 100  & 86.7  & 89.9 \\ \midrule

MOJITO (Ours)           & C \& L & 0      &     \textbf{98.6} &  \textbf{96.9}    &  94.5    & \textbf{100}      &   \textbf{83.5}   & \textbf{\textbf{88.9}} \\  \bottomrule
\end{tabular}
\label{table_quantitative_comparsion_on_navsim_v1}
\end{table}

% Table\ref{table_quantitative_comparsion\subsection{Quantitative Comparison}
To evaluate our proposed method, we compare it against two distinct categories of autonomous driving frameworks on the NAVSIM-v1 dataset: End-to-End methods and  Vision-Language-Action (VLA) models. 

\textbf{Comparison with End-to-End Methods.} 
Table~\ref{table_quantitative_comparsion_on_navsim_v1} presents the comparison with representative end-to-end methods. 
Overall, our method achieves a leading PDMS score of 88.9, outperforming recent baselines such as DiffusionDrive (88.1 PDMS) and WoTE (88.3 PDMS). Many  high-performing methods heavily rely on predefined anchors to constrain the planning space, such as VADv2 and Hydra-MDP (8,192 anchors), WoTE (256 anchors), and DiffusionDrive (20 anchors). In contrast, our method completely eliminates this requirement. Despite using zero anchors, our approach surpasses DiffusionDrive, VADv2, and Hydra-MDP. This indicates that our architecture can generate superior and stable actions without being restricted by predefined anchors.

\begin{table}[!t]
\caption{Quantitative comparison with VLA-based methods on NAVSIM-v1 (\textit{navtest}) using non-reactive closed-loop evaluation. ``VLM'' denotes the vision-language model backbone used by each method, ``RL'' indicates whether reinforcement learning finetuning is applied, and ``Model Size'' reports the parameter count. }
\renewcommand{\arraystretch}{0.8}
\setlength{\tabcolsep}{3.2mm}\begin{tabular}{l|lccc}
\hline \toprule
Method              & VLM        & RL        & Model Size & PDMS $\uparrow$\\ \midrule
ReCogDrive-Base-IL~\cite{li2025recogdrive}  & InternVL3  &           & 2B         & 86.5 \\
ReCogDrive-Base-RL~\cite{li2025recogdrive}  & InternVL3  & \ding{51} & 2B         & \textbf{90.8} \\
ReCogDrive-Large-IL~\cite{li2025recogdrive} & InternVL3  &           & 8B         & 86.5 \\
ReCogDrive-Large-RL~\cite{li2025recogdrive} & InternVL3  & \ding{51} & 8B         & 90.4 \\ \midrule
AutoVLA-IL~\cite{zhou2025autovla}          & Qwen2.5-VL &           & 3B         & 80.5 \\
AutoVLA-RL~\cite{zhou2025autovla}          & Qwen2.5-VL & \ding{51} & 3B         & 89.1 \\ \midrule
AdaThinkDrive-IL~\cite{luo2025adathinkdrive}    & InternVL3  &           & 8B         & 86.2 \\
AdaThinkDrive-RL~\cite{luo2025adathinkdrive}    & InternVL3  & \ding{51} & 8B         & 90.3 \\ \midrule
MOJITO (Ours)                & None       &           & 127M       & 88.9 \\ \bottomrule
\end{tabular}
\vspace{-1.5em}
\label{table_quantitative_comparsion_on_navsim_v1 VLM}
\end{table}

\textbf{Comparison with VLA-based Methods.} 
Recently, VLA models have introduced large foundation models to autonomous driving. As shown in Table~\ref{table_quantitative_comparsion_on_navsim_v1 VLM}, we compare with representative VLA frameworks  built upon large foundation models like InternVL3 and Qwen2.5-VL. A notable characteristic of these VLA methods is their massive model sizes, ranging from 2B to 8B parameters. To achieve competitive driving performance, they heavily rely on Reinforcement Learning (RL) techniques. For instance, ReCogDrive-Base requires RL to improve its PDMS from 86.5 to 90.8.

In contrast, our method operates with a compact model size of 127M parameters. Without utilizing  Vision-Language Models or complex Reinforcement Learning pipelines, our approach achieves a PDMS of 88.9, achieving a competitive score compared to VLM-RL-based methods. This demonstrates the exceptional parameter efficiency and learning capability of our proposed architecture.

\begin{table}[t]
\caption{Quantitative comparison on NAVSIM-v2 \textit{navtest} under non-reactive closed-loop evaluation. NAVSIM-v2 reports extended compliance metrics (e.g., DDC, TLC, LK, HC, and EC) in addition to the v1 metrics, and uses EPDMS as the overall score. }
%\renewcommand{\arraystretch}{0.95}
% \setlength{\tabcolsep}{0.6mm}
% \resizebox{0.97\textwidth}{!}{
% \begin{tabular}{l|ccccccccc|c}
% \hline \toprule
% Method & NC $\uparrow$ & DAC $\uparrow$ & DDC $\uparrow$ & TLC $\uparrow$ & EP $\uparrow$ & TTC $\uparrow$ & LK $\uparrow$ & HC $\uparrow$ & EC $\uparrow$ &  EPDMS $\uparrow$ \\ \midrule
% Ego Status MLP          & 93.1 & 77.9 & 92.7 & 99.6 & 86.0 & 91.5 & 89.4 & \ul{98.3} & 85.4 &  64.0 \\
% Transfuser~\cite{chitta2022transfuser}              & 96.9 & 89.9 & 97.8 & 99.7 & 87.1 & 95.4 & 92.7 & \ul{98.3} & 87.2 &  76.7 \\
% Hydra-MDP++~\cite{li2024hydra}             & 97.2 & \textbf{97.5} & \ul{99.4} & 99.6 & 83.1 & 96.5 & 94.4 & 98.2 & 70.9 &  81.4 \\
% DriveSuprim~\cite{yao2025drivesuprim}             & 97.5 & 96.5 & \ul{99.4} & 99.6 & \ul{88.4} & 96.6 & 95.5 & \ul{98.3} & 77.0 &  83.1 \\
% ARTEMIS~\cite{feng2025artemis}                & \textbf{98.3} & 95.1 & 98.6 & \textbf{99.8} & 81.5 & \textbf{97.4} & \ul{96.5} & \ul{98.3} & -    &  83.1 \\ 
% %MindDrive~\cite{sun2025minddrive}              & 98.4 & 95.6 & 99.3 & 99.8 & 86.7 &97.5  & 96.5 & 100 & 96.8 & 84.2 \\ 
% DiffusionDriveV2~\cite{zou2025diffusiondrivev2}       & 97.7 & \ul{96.6} & 99.2 & \textbf{99.8} & \textbf{88.9} & 97.2 & 96.0 & 97.8 & 91.0 &  85.5 \\ \midrule

% MOJITO (Ours)                    & \ul{98.2} & 96.2 & \textbf{99.5} & \textbf{99.8} & 87.8 & \textbf{97.4} & \textbf{97.3} & \textbf{98.4} & \ul{87.8} &  \textbf{88.4} \\ \bottomrule

% \end{tabular}}
% \label{table_quantitative_comparison}
% \end{table}

\setlength{\tabcolsep}{0.6mm}
\resizebox{0.99\textwidth}{!}{
\begin{tabular}{l|ccccccccc|c}
\hline \toprule
Method & NC $\uparrow$ & DAC $\uparrow$ & DDC $\uparrow$ & TLC $\uparrow$ & EP $\uparrow$ & TTC $\uparrow$ & LK $\uparrow$ & HC $\uparrow$ & EC $\uparrow$ &  EPDMS $\uparrow$ \\ \midrule
Ego Status MLP          & 93.1 & 77.9 & 92.7 & 99.6 & 86.0 & 91.5 & 89.4 & 98.3 & 85.4 &  64.0 \\
Transfuser~\cite{chitta2022transfuser}              & 96.9 & 89.9 & 97.8 & 99.7 & 87.1 & 95.4 & 92.7 & 98.3 & 87.2 &  76.7 \\
Hydra-MDP++~\cite{li2024hydra}             & 97.2 & \textbf{97.5} & 99.4 & 99.6 & 83.1 & 96.5 & 94.4 & 98.2 & 70.9 &  81.4 \\
DriveSuprim~\cite{yao2025drivesuprim}             & 97.5 & 96.5 & 99.4 & 99.6 & 88.4 & 96.6 & 95.5 & 98.3 & 77.0 &  83.1 \\
ARTEMIS~\cite{feng2025artemis}                & \textbf{98.3} & 95.1 & 98.6 & \textbf{99.8} & 81.5 & \textbf{97.4} & 96.5 & 98.3 & -    &  83.1 \\ 
%MindDrive~\cite{sun2025minddrive}              & 98.4 & 95.6 & 99.3 & 99.8 & 86.7 &97.5  & 96.5 & 100 & 96.8 & 84.2 \\ 
DiffusionDriveV2~\cite{zou2025diffusiondrivev2}       & 97.7 & 96.6 & 99.2 & \textbf{99.8} & \textbf{88.9} & 97.2 & 96.0 & 97.8 & \textbf{91.0} &  85.5 \\ \midrule

MOJITO (Ours)                    & 98.2 & 96.2 & \textbf{99.5} & \textbf{99.8} & 87.8 & \textbf{97.4} & \textbf{97.3} & \textbf{98.4} & 87.8 &  \textbf{88.4} \\ \bottomrule

\end{tabular}}
\label{table_quantitative_comparison}
\end{table}
%\begin{table*}[htbp]
\begin{table*}[t]
\caption{Results on NAVSIM-v2 \textit{navhard} under the two-stage evaluation protocol. We report Stage I and Stage II metrics and the overall EPDMS. Since \textit{navhard} does not provide LiDAR for Stage II, we evaluate MOJITO under a camera-only setting.}
  \centering 
  \renewcommand{\arraystretch}{1.1}
  \resizebox{0.97\textwidth}{!}{
  \begin{tabular}{l|l|ccccccccc|c}
    \toprule
    Method & Stage & NC$\uparrow$ & DAC$\uparrow$ & DDC$\uparrow$ & TLC$\uparrow$ & EP$\uparrow$ & TTC$\uparrow$ & LK$\uparrow$ & HC$\uparrow$ & EC$\uparrow$ & EPDMS$\uparrow$ \\
    \midrule
    \multirow{2}{*}{LTF~\cite{chitta2022transfuser}} & Stage I & 96.2 & 79.5 & {99.1} & {99.5} & 83.1 & 95.1 & 94.2 & 97.5 & {79.1} & \multirow{2}{*}{23.1} \\
    & Stage II & 77.7 & 70.2 & 84.2 & 98.0 & 85.1 & 75.6 & 45.4 & 95.7 & {75.9} & \\
    \midrule
    \multirow{2}{*}{DiffusionDrive~\cite{liao2025diffusiondrive}} & Stage I & 96.0 & 79.7 & 97.4 & {99.5} & 81.3 & 93.1 & 90.8 & 96.8 & 73.8 & \multirow{2}{*}{24.2} \\
    & Stage II & 82.1 & 72.2 & 88.5 & 98.7 & 85.1 & 78.8 & 49.2 & 89.3 & 71.2 & \\
    \midrule
    \multirow{2}{*}{GTRS-DP~\cite{li2025gtrs}} & Stage I & 94.7 & 78.8 & 96.1 & {99.5} & 83.0 & 94.4 & 92.0 & 97.5 & 72.8 & \multirow{2}{*}{23.8} \\
    & Stage II & 80.3 & 74.4 & 84.9 & 98.0 & 81.9 & 78.8 & 45.4 & 96.7 & 70.1 & \\
    \midrule
    \multirow{2}{*}{GuideFlow~\cite{liu2025guideflow}} & Stage I & {96.6} & 80.5 & 96.3 & 99.3 & 82.3 & 94.9 & 91.5 & {97.7} & 67.8 & \multirow{2}{*}{27.1} \\
    & Stage II & {87.3} & 76.7 & {88.8} & {99.2} & 84.3 & {85.1} & {49.7} & 93.1 & 44.5 & \\
    \midrule
    % \multirow{2}{*}{{MindDrive}} & Stage I & 96.1 & {86.0} & 98.8 & 99.3 & {83.3} & {95.6} & {94.4} & 97.6 & 74.7 & \multirow{2}{*}{{30.5}} \\
    % & Stage II & 82.6 & {79.1} & 86.4 & 98.0 & {85.3} & 79.4 & 49.2 & 96.5 & {71.0} & \\
    % \midrule
    \multirow{2}{*}{\thead{MOJITO \\ Camera-Only}} & Stage I & 96.2 & 82.4 & 98.1 & 99.6 & 84.1 & 94.7 & 94.2 & 97.6 & 80.0 & \textbf{\multirow{2}{*}{29.0}} \\
    & Stage II & 78.0 & 69.4 & 85.0 & 97.9 & 85.5 & 75.4 & 48.5 & 96.0 & 71.6 & \\
    \bottomrule
  \end{tabular}}
  \label{table:navhard_table}
\end{table*}
\subsection{Quantitative Comparison on NAVSIM-v2}

We further test our method under the  stricter NAVSIM-v2 protocol using the same \textit{navtest} split. 
% While NAVSIM-v1 focuses primarily on overall driving quality, NAVSIM-v2 introduces a richer, expanded set of compliance metrics (such as Driving Direction Compliance, Lane Keeping, and Ego Compliance) to enable a fine-grained examination of planner behavior.
Because the \textit{navtest} split lacks  synthetic follow-up observations required for Stage-2 causal evaluation, we focus our comparison on the Stage-1 overall score (EPDMS).
Table~\ref{table_quantitative_comparison} summarizes the results. Our method achieves the best EPDMS of 88.4, outperforming DiffusionDriveV2 (85.5), ARTE-MIS (83.1), and DriveSuprim (83.1). It also performs strongly on key compliance metrics, with 99.5 DDC, 97.3 LK, and 98.4 HC.

We additionally report results on NAVSIM-v2 \textit{navhard}, which supports both Stage-1 and Stage-2 evaluation. 
% As \textit{navhard} does not provide LiDAR for Stage-2, we evaluate our method under camera-only setting. 
As \textit{navhard} does not provide LiDAR for Stage-2, we remove the LiDAR branch of our model and evaluate under a camera-only setting. This causes a performance drop compared to using both modalities, but the results remain strong overall.
As shown in Table~\ref{table:navhard_table}, our method reaches 29.0 EPDMS, exceeding GuideFlow (27.1), DiffusionDrive (24.2), and LTF (23.1). Stage-2 is consistently harder than Stage-1 for all methods under this setting.

\subsection{Qualitative Comparison on NAVSIM-v1}
\begin{figure}[tbp]
\centering
\includegraphics[width=\linewidth]{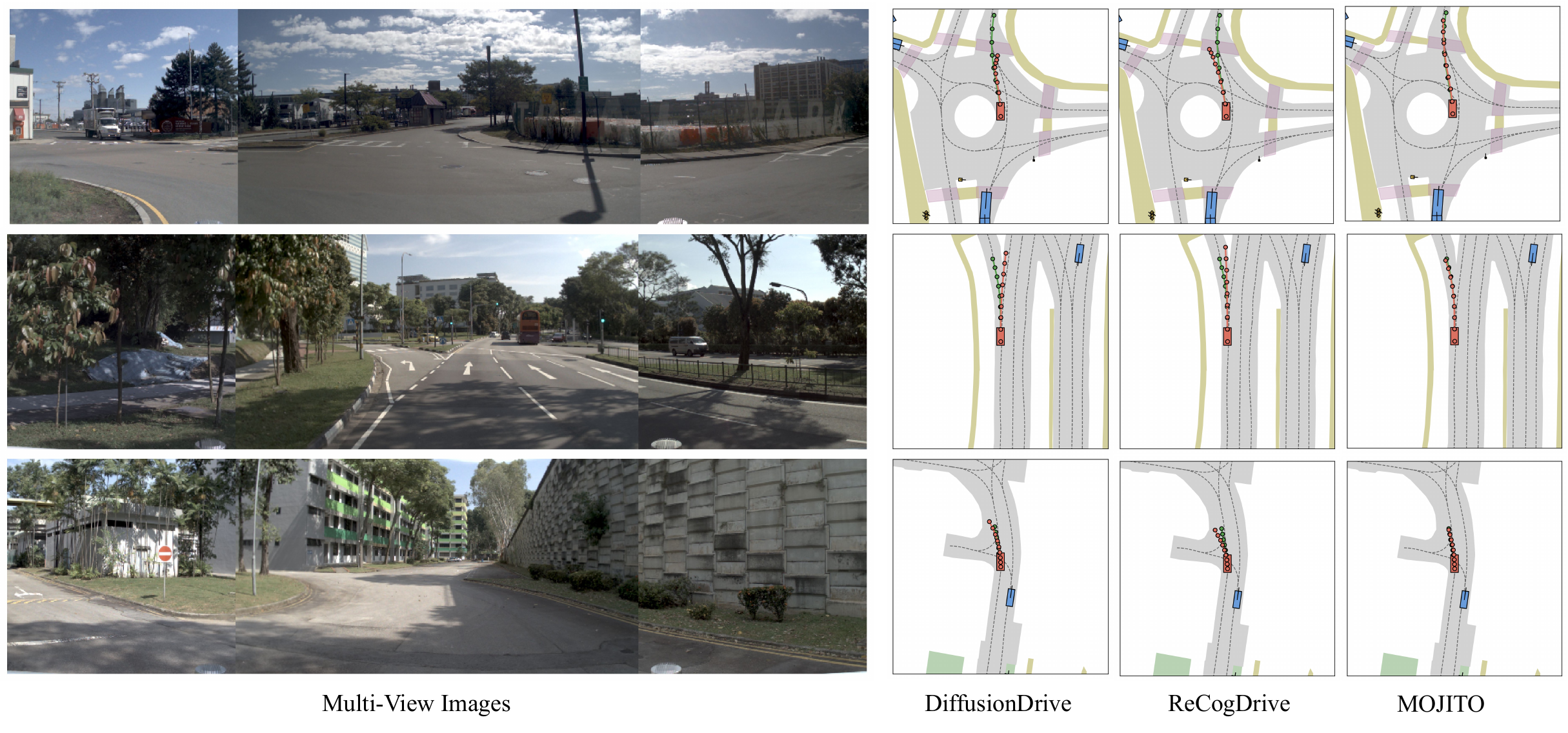}
\caption{This figure presents a qualitative comparison between our method and existing methods on NAVSIM-v1. Our proposed MOJITO achieves more stable and reliable decision-making in complex scenarios. The green  line denotes the ground-truth trajectory,
while the red  line represents the trajectory generated by DiffusionDrive, ReCogDrive, and MOJITO, respectively.
}
\vspace{-1em}
\label{qualitative comparsion}
\end{figure}

To  demonstrate the superiority of our proposed MOJITO, we conduct a qualitative comparison on NAVSIM-v1, as shown in Fig. \ref{qualitative comparsion}.  The first row illustrates a challenging roundabout scenario with multiple drivable routes.  MOJITO generates a  precise, smooth, and stable trajectory.
The second row depicts a road bifurcation with slight directional differences,  and MOJITO successfully distinguishes the correct lane.  The third row shows that other methods tend to produce overly aggressive turning behaviors, potentially driving into the undrivable area. MOJITO achieves a stronger understanding of scene geometry and depth, enabling it to generate more stable and reliable trajectories.

\subsection{Ablation Study}
\begin{table}[t]
\caption{Ablations on sensor modalities, LiDAR tokenization, and the attention formulation on NAVSIM-v1 \textit{navtest}.}
\renewcommand{\arraystretch}{0.9}
\resizebox{0.99\textwidth}{!}{
\setlength{\tabcolsep}{1.3mm}\begin{tabular}{c|cccc|ccccccc}
\toprule
ID & Image         & LiDAR         & Sample & Attn.      & NC$\uparrow$   & DAC$\uparrow$  & TTC  $\uparrow$& Comf.$\uparrow$ & EP $\uparrow$  & PDMS $\uparrow$\\ \midrule
1  & \ding{51}     & \ding{55}     & \ding{55}       & Self-Attn.  & 98.0 & 95.3 & 93.6 &  \textbf{100}    & 81.6 & 86.8 \\
2  & \ding{51}     & \ding{51}     & FPS+KNN    & Self-Attn.  & 97.8 & 94.8 & 93.3 &  \textbf{100}    & 81.0 & 86.1 \\
3  & \ding{51}     & \ding{51}     & Pillar & Cross-Attn. &  97.8    &   94.4   &   93.1   &  \textbf{100}     & 80.7     &    85.7\\
4  & \ding{51}     & \ding{51}     & Pillar & Self-Attn.  & \textbf{98.6} & \textbf{96.9} & \textbf{94.5} & \textbf{100}   & \textbf{83.5} & \textbf{88.9} \\ \bottomrule
\end{tabular}
}
\label{tab:ablation}
\end{table}
\begin{table}[t]
\caption{Scaling study on NAVSIM-v1 (\textit{navtest}). We increase the aligned blocks in the vision, LiDAR, and action planner branches, initializing the perception branches from the corresponding pretrained blocks. The best results are marked in bold.}
% \vspace{-1em}
\renewcommand{\arraystretch}{0.9}
\setlength{\tabcolsep}{2.6mm}

\resizebox{\textwidth}{!}{%

\begin{tabular}{l|cc|cccccc}
\hline \toprule
ID & Blocks & Model Size & NC$\uparrow$   & DAC$\uparrow$  & TTC$\uparrow$  & Comf.$\uparrow$ & EP  $\uparrow$ & PDMS $\uparrow$\\ \midrule
1  & 3      & 33.4M      &  96.6    &90.4      &90.7     &\textbf{100}       &76.1      & 80.5 \\
2  & 6      & 64.4M      & 96.7 & 90.2 &90.7  &\textbf{100}   &76.1  & 80.5 \\
3  & 9      & 95.5M      & 97.1 & 92.3 & 91.8 & \textbf{100}  & 78.4 & 83.0 \\
4  & 12     & 127M       & \textbf{98.6} & \textbf{96.9} & \textbf{94.5} & \textbf{100}   & \textbf{83.5} & \textbf{88.9} \\ \bottomrule
\end{tabular}
}
\vspace{-1em}
\label{table_scalability}
\end{table}
% Table~\ref{tab:ablation} reports ablations on NAVSIM \textit{navtest}. We study three factors: sensor modalities (camera vs. camera+LiDAR), LiDAR tokenization (FPS+KNN vs. Pillar-based grouping), and the attention formulation in our modal joint attention blocks (cross-attention vs. self-attention). PDMS is used as the primary metric.
%Table~\ref{tab:ablation} presents ablations on NAVSIM \textit{navtest}. We vary the input modalities (camera vs.\ camera+LiDAR), LiDAR tokenization (FPS+KNN vs.\ PillarGroup), and the attention type in our Modal Joint Attention blocks (self-attn.\ vs.\ cross-attn.). 
In this section, we conduct ablation experiments on the NAVSIM-v1 navtest split, as reported in Table~\ref{tab:ablation}. We vary the input sensor modalities, the LiDAR tokenization strategy, and the attention type in our Modal Joint Attention blocks. These ablations provide insights into the effectiveness of our unified architecture and the necessity of each design choice.

ID 1 evaluates a camera-only variant of our model, which still achieves a competitive PDMS of 86.8, indicating that directly fusing trajectory tokens with image tokens provides strong planning guidance. In ID 2, we incorporate LiDAR inputs using an  FPS+KNN tokenization strategy, which is commonly adopted in shape-level point cloud representation learning. This tokenization  does not improve performance, and we attribute this degradation to the incompatibility with large-scale autonomous driving scenes.
%Replacing FPS+KNN tokenization with our Pillar-based grouping improves the effectiveness of LiDAR features (ID2 vs. ID4), highlighting the importance of LiDAR tokenization for driving scenarios. 
% Self-attention enables bidirectional interaction among trajectory, image, and LiDAR tokens, so sensor tokens can be adaptively re-weighted conditioned on the evolving plan.

In ID 3, we prevent the sensor tokens from attending to the action tokens by modifying the attention map within the Joint Modal Attention. Cross-attention allows action tokens to understand sensor inputs,  while the sensor tokens do not adapt to the current denoising state.  Our self-attention mechanism fuses image, LiDAR, and action tokens simultaneously, enabling the bidirectional interaction between  perception and  action planning.
% Self-attention lets trajectory, image, and LiDAR tokens attend to each other, so sensor tokens are updated along with the evolving plan. Stacking these blocks lets the planner refine the trajectory step by step, improving overall planning quality.
% When stacked over multiple blocks, this layer-wise exchange facilitates iterative refinement during trajectory generation and leads to better overall planning quality.

\subsection{Scalability Study}

Table~\ref{table_scalability} studies how performance scales with model depth. The ``Blocks'' denote aligned blocks that are increased jointly in the image, LiDAR, and action planner branches. For the image and LiDAR branches, we load the corresponding pre-trained weights for the selected number of blocks and  train the model in an end-to-end manner. Even the smallest model variant achieves reasonable performance, and scaling up the number of blocks consistently improves PDMS. 
This trend indicates that our architecture is easy to scale in model size and can benefit from larger capacity, suggesting further gains when trained on larger-scale data. Moreover, we also train the action planner from scratch for a fairer comparison, and the scaling trend remains consistent.

\subsection{Instruction-Following and Diverse Trajectory Generation}
\begin{figure}[t]
\centering
\includegraphics[width=0.9\linewidth]{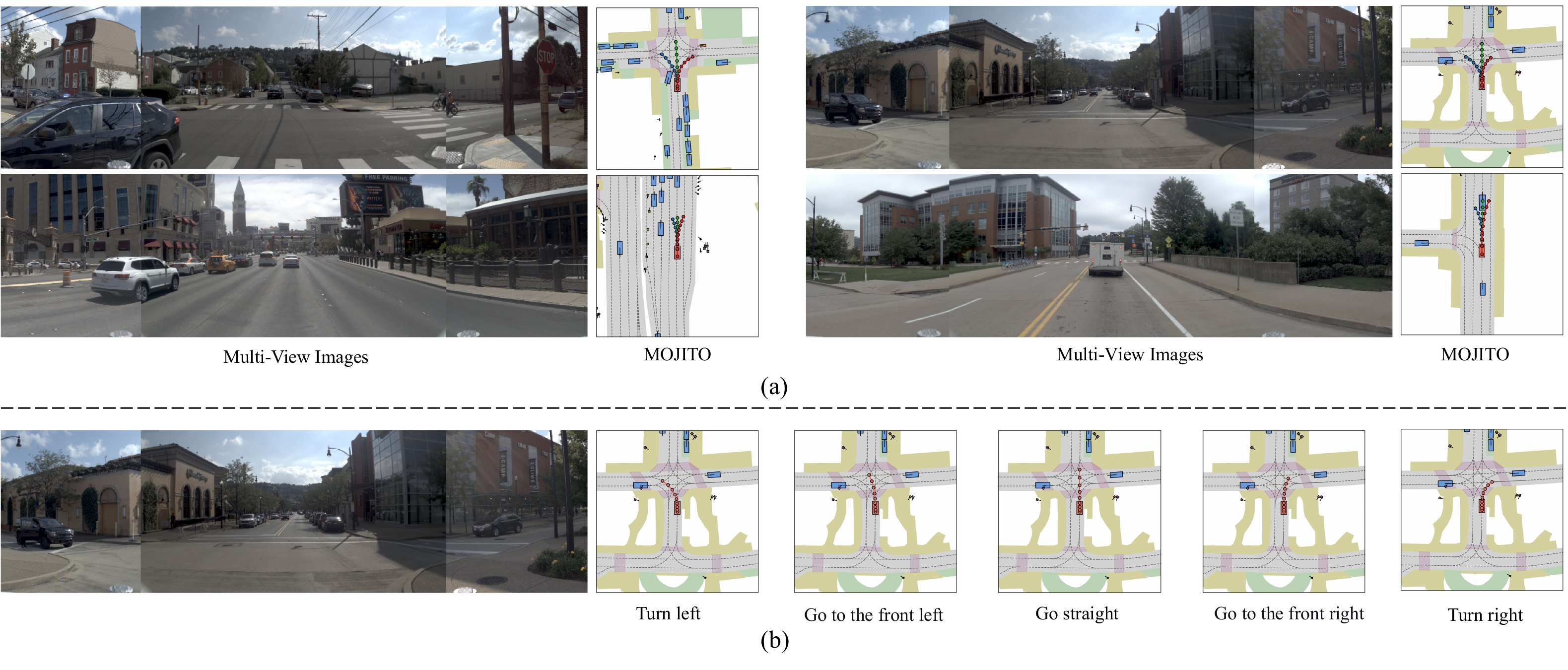}
\caption{
This figure demonstrates the instruction-following  and diverse trajectory generation capability of MOJITO. (a) The blue, green, and red lines represent the trajectories generated for the instructions ``turn left'', ``go straight'', and ``turn right'', respectively.
%The first row shows the scenario at an intersection, while the second row illustrates a straight-lane scenario. 
(b) shows our ability to follow more complex instructions. Benefiting from
our diffusion-based planner, our method can generate trajectories with varying turning
angles, further validating the diversity of its trajectory generation.
}
\vspace{-1em}
\label{language}
\end{figure}

VLM-based driving methods often emphasize instruction-following through language understanding~\cite{yang2024diffusiones, zhou2025autovla}.  MOJITO does not rely on a VLM backbone, but it  supports instruction-following via a high-level command interface. In our examples, we construct a command vocabulary and use Qwen3-1.7B~\cite{qwen3} to map  language instructions to the closest commands, which are then fed into the model as the conditions. As shown in Fig.~\ref{language} (a), MOJITO accurately follows the specified command and produces lane-consistent trajectories toward the intended direction. 
The second
row in Fig.~\ref{language} (a) shows that in the straight-lane scenario, even with left or
right turn commands, MOJITO does not generate unreasonable sharp turns.
%The second row in Fig.~\ref{language} (a) shows that in the straight-lane scenario, even with  left or right turn commands, MOJITO does not generate unreasonable sharp turns. Moreover, 
As a diffusion-based planner, MOJITO  enables diverse and smooth trajectory generation in a continuous noise space. As shown in Fig.~\ref{language} (b), as the turning angle implied by the instruction gradually varies, MOJITO consistently generates feasible  and stable trajectories.

\begin{figure}[t]
\centering
\includegraphics[width=0.9\linewidth]{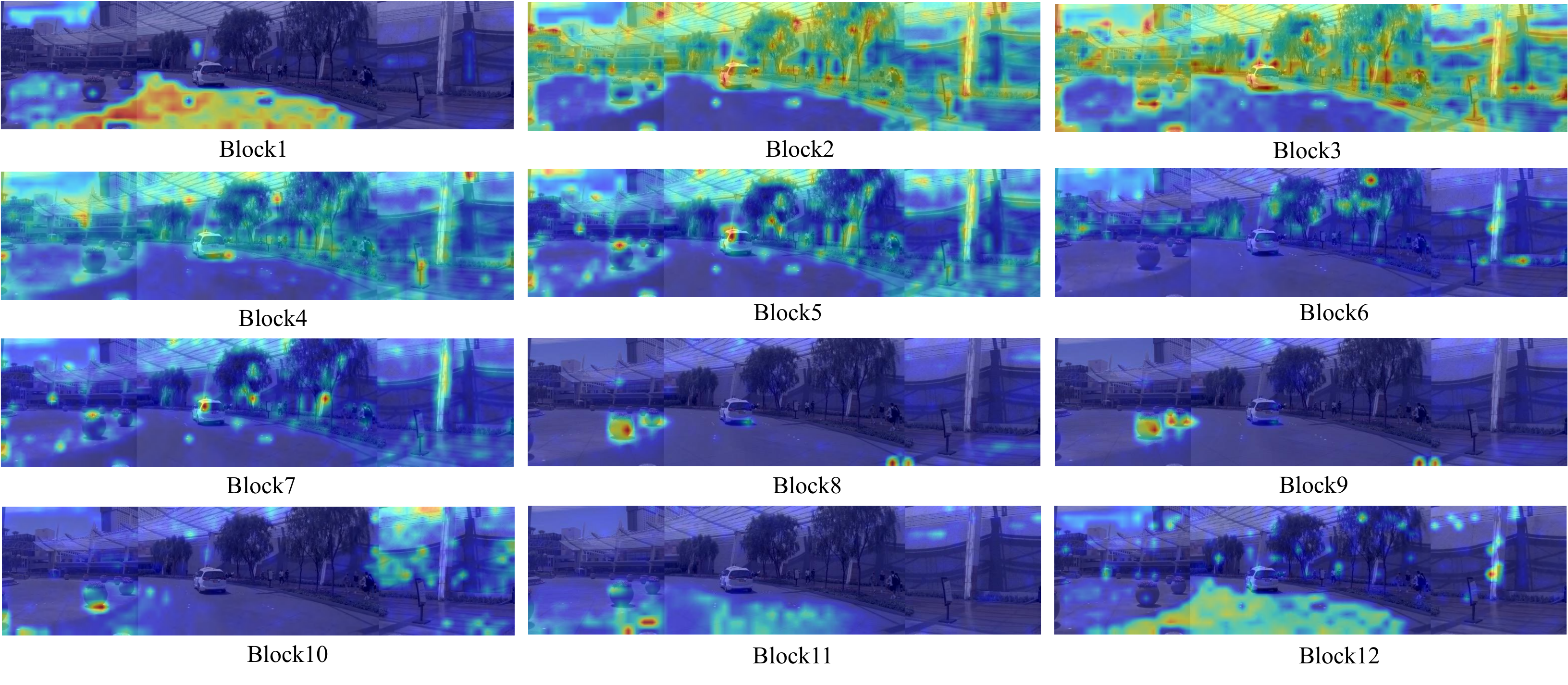}
% \vspace{-1em}
\caption{Visualization of the trajectory-to-image attention maps. The attention weights are averaged over the 8 future waypoints to produce a holistic view of how the planning branch attends to the multi-view image patches.}
\vspace{-1em}
\label{attn_cr}
\end{figure}

\section{Cross Modal Learning}
To enable the planning branch to effectively exploit visual semantics, we introduce the  Modal Joint Attention that facilitates direct interaction between the action planner and the image branch.  To better understand this interaction, we visualize the attention maps derived from the Modal Joint Attention across different transformer blocks. As shown in Fig.~\ref{attn_cr},  the first block focuses on the forward drivable area; the middle blocks shift to large foreground objects and obstacles on both road sides; and the last blocks exhibit fine-grained attention to both the drivable area and roadside obstacles. This attention pattern validates that the planner dynamically adapts its visual attention to the driving context, contributing to more stable and precise trajectory planning.

\section{Inference Latency}
In this section, we measure per-sample latency on a single H200 GPU using BF16 precision.  MOJITO employs 2 diffusion steps  and achieves an inference latency of 187.65\,ms, while Transfuser~\cite{chitta2022transfuser}, DiffusionDrive~\cite{liao2025diffusiondrive}, and ReCogDrive~\cite{li2025recogdrive} have latencies of 88.15\,ms, 123.22\,ms, and 331.68\,ms, respectively. Although our inference speed is slightly slower than methods such as Transfuser and DiffusionDrive, MOJITO significantly outperforms them on the NAVSIM-v1 and NAVSIM-v2 datasets. In practical autonomous driving scenarios, the safety and stability of driving decisions are of critical importance. Moreover, compared to VLA-based methods like ReCogDrive, MOJITO achieves much faster inference while maintaining   superior performance. These results demonstrate that our method strikes a good balance between inference efficiency and trajectory accuracy. 

\section{Conclusion}
In this work, we presented \textbf{MOJITO}, a unified end-to-end driving framework that avoids the  information bottleneck induced by the cascaded pipelines. 
MOJITO adopts three parallel Transformer branches for image, LiDAR, and action planning, and introduces a block-wise Modal Joint Attention to fuse all modalities simultaneously.  The action planner branch can  leverage fine-grained sensor inputs without auxiliary supervision or predefined trajectory anchors. Extensive experiments on NAVSIM validate the effectiveness of our design, achieving 88.9 PDMS on NAVSIM-v1 \textit{navtest} and 88.4 EPDMS on NAVSIM-v2 \textit{navtest}.
\section*{Acknowledgments}
This work was supported by the National Natural Science Foundation of China (NSFC) under Grant U22A2094 and Grant 62272435. 

\bibliographystyle{splncs04}
\bibliography{main}
\end{document}